# EVALUATING AND VALIDATING CLUSTER RESULTS

Anupriya Vysala and Dr. Joseph Gomes

Department of Computer Science, Bowie State University, USA

## ABSTRACT

*Clustering is the technique to partition data according to their characteristics. Data that are similar in nature belong to the same cluster [1]. There are two types of evaluation methods to evaluate clustering quality. One is an external evaluation where the truth labels in the data sets are known in advance and the other is internal evaluation in which the evaluation is done with data set itself without true labels. In this paper, both external evaluation and internal evaluation are performed on the cluster results of the IRIS dataset. In the case of external evaluation Homogeneity, Correctness and V-measure scores are calculated for the dataset. For internal performance measures, the Silhouette Index and Sum of Square Errors are used. These internal performance measures along with the dendrogram (graphical tool from hierarchical Clustering) are used first to validate the number of clusters. Finally, as a statistical tool, we used the frequency distribution method to compare and provide a visual representation of the distribution of observations within a clustering result and the original data.*

## 1. INTRODUCTION

Clustering is grouping sets of data objects into multiple groups or clusters so that objects within the cluster have high similarity, but are very dissimilar to the objects in the other clusters [2]. The clustering algorithms can be used on both normalized and non-normalized data. If users have normalized data, the number of iteration of the algorithms will be fewer. Therefore, in most of the situation, normalized data offers good outcome as compared to non-normalized data[3].In machine learning, there are both supervised and unsupervised learning algorithms. In the case of supervised learning, we feed training data along with predefined labels, whereas in the case of unsupervised learning we only feed features when we train data as the true label is not known. In general, unsupervised algorithms make inferences from datasets using only input vectors without referring to the labeled outcomes. K-means is one of the Unsupervised learning algorithms. There are many types of clustering such as exclusive, overlapping, hierarchical, etc. In the case of exclusive clustering, each sample in the dataset will belong to one of the clusters, whereas in the case of overlapping clusters there can be data points that are in more than one cluster. In the case of hierarchical clustering,a tree-like structure is produced.

In this paper, we first validate the number of clusters by variousinternal evaluation techniques and then compare those results with the K-means elbow method and hierarchical agglomerative clustering. Later we do the cluster analysis based on internal evaluation techniques and frequency distribution method for both K-means and Hierarchical Clustering Algorithm.We have implemented these in Python.

The rest of the paper is organized as follows: Section 2 details about K-means and Hierarchical Clustering, Section 3 offers surveys of related works. Section 4 offers various external and internal evaluation methods. Section 5 explains the proposed methodology. Section 6discussesthe findings from these results. Finally, we conclude by briefly explaining our contributions and further works.

## 2. K-MEANS AND HIERARCHICAL CLUSTERING

### 2.1. K-means

K-means is one of the popular unsupervised machine learning algorithms for clustering. K-means is fast, compared to other clustering algorithms as it doesn't require calculating all of the distances between each observation and every other observation. K-means can be used for clustering very large datasets. But traditional k means algorithm does not always generate good quality results as automatic initialization of centroids affects final clusters as mentioned in [4].

In this paper, we used the k-means algorithm on a normalized dataset in order to generate quality clusters, it minimized the k-means inertia criteria (within-cluster sum of square inertia) and gave high score in case of various external evaluation techniques, which can be seen in section 6, Table 1.

The k-means model was implemented using the scikit-learn library in python where we used the k-means++ for initialization as 'k-means++' selects initial cluster centers for k-means clustering in a smart way to speed up convergence. In our experiment, the training data was split into random train and test subsets using 80-20 rule and the random state was set to an integer value to have a reproducible result for documenting, in order to avoid different solutions every time, we do sampling.

### 2.1.1. Algorithm (K-means)[2]

The algorithm proceeds as follows:
1. Randomly initialize'k' clusters centroids
2. Assign each data points to its closest cluster centroid
3. Compute the centroid of the new partition formed by taking the average of points assigned to that cluster
4. Repeat steps 2, and 3 until convergence is obtained

### 2.2. Hierarchical Clustering

Another type of clustering technique is called Hierarchical Clustering. The two types of Hierarchical clustering are Agglomerative clustering and Divisive clustering. In the case of hierarchical agglomerative clustering it begins by putting each observation into its own separate cluster and then it builds new higher-level clusters by grouping clusters based on distance. The graphical tool to get insight into the clustering solution is called a dendrogram.

In hierarchical clustering, as a first step, find the distance matrix and then examine all the distances between all the observations and pairs together the two closest ones to form a new cluster. So finding the first cluster simply means looking for the lowest number in the distance matrix and merging the two observations that the distance corresponds to into a new cluster. Hence there will be one fewer cluster than the number of observations. To determine which observations will form the next cluster certain linkage methods are used. The different linkage methods used are single linkage, complete linkage, average linkage, and ward.

The Single linkage method defines cluster proximity as the shortest distance (min) between two points, $x$, and $y$, that are in different clusters, $A$ and $B$. Complete linkage defines cluster proximity as the farthest(max) distance between two points, $x$ and $y$, that are in different clusters, $A$ and $B$. Average proximity defines cluster proximity as the average distance between two points, $x$ and $y$, that are in different clusters, $A$ and $B$. In ward's method, the approach of calculating the similarity between two clusters is exactly the same as group average except that Ward's method calculates the sum of the square of the distances 'Pi' and 'Pj' where 'Pi' is a data point in cluster 1 and 'Pj' is a data point in cluster 2. In our paper, we have compared the results using all these methods for the IRIS dataset and figured out which one gives a better score.

### 2.2.1. Algorithm (Hierarchical Clustering) [5]

1. Start with n clusters where each data object forms a single cluster
2. Repeat step 2 until there is only one cluster left
   (a) Find the closest pair of clusters
   (b) Merge them
3. Return the tree of cluster-merger

## 3. RELATED WORK

The paper [1] does a comparison of K-Means Clustering and CLARA Clustering on Iris Dataset, which uses Euclidean distance and Manhattan Distance as a dissimilarity measure, respectively. It is an extension of the K-Medoids Clustering algorithm, which uses a sampling approach to handle large datasets. The paper proves that CLARA Clustering using Manhattan distance is better than K-Means Clustering with Euclidean distance.

Dr. Manju Kaushik et al. [2] presented a comparative study of K-means and Hierarchical Clustering techniques in terms of Clustering Criteria, performance, size of a dataset, sensitivity to noise, quality, execution time, number of clusters, etc.

In [4] Deepali Virmani et al., have first preprocessed the dataset based on the normalization technique and then generated effective clusters. They also assigned a weight to each attribute value to ensure the fair distribution of clusters. Their algorithm has proved to be better than the traditional K-means algorithm in terms of execution time and speed.

Weka data mining tool version 3.7 was used in [5] for testing accuracy and running time of k-means and hierarchical clustering algorithms on the IRIS and Diabetes dataset. In this paper, they concluded that the accuracy of k-means for the IRIS dataset having 'real' datatype attributes was greater compared to hierarchical clustering and for the diabetes dataset having both integer and real attributes accuracy of hierarchal clustering was greater than the k-means clustering algorithm. In the case of running time, k-means was faster.

Rousseeuw [6] proposed a technique called silhouette analysis which is based on the comparison of object tightness and separation. This tool can be used to validate the number of clusters in the case of the clustering algorithm.

Kalpit et al., [7] compared K-means and K-medoids algorithms using the Iris dataset. The results showed that K-medoids is better than K-means at scalability for a larger dataset. K-medoids showed its superiority over k means in execution time and sensitivity towards outliers.

In [9], Tippaya Thinsungnoena et al. studied two interesting sub-activity of the clustering process -selecting the number of clusters and analyzing the result of data clustering using SSE and silhouette analysis.

In [11] Huda et al. studied the several applications of the k-means algorithm in data mining and pattern recognition. They showed that k-means produces a promising result in both as a clustering method in data mining and for segmenting images in pattern recognition. Thus the paper concluded that k-means clustering is an efficient algorithm in both areas.

T. Velmurugan et.al., from the experiments they performed, concluded in [13] that K-Medoids is more robust than K-Means clustering in terms of noise and outliers, although K-Medoids is good for only small datasets.

## 4. VARIOUS EXTERNAL AND INTERNAL EVALUATION METHODS

### 4.1. External Evaluation

**Homogeneity:** If each cluster in the clustering result contains only data points that are members of a single class, then it satisfies homogeneity.

**Completeness:** If all the data points that are members of a given class are elements of the same cluster in the clustering result, then it satisfies completeness.

**V- measure:** The V-Measure is defined as the harmonic mean of homogeneity and completeness of the clustering.

### 4.2. Internal Evaluation

#### 4.2.1. Silhouette Analysis

According to Rousseeuw [6], Silhouette is a tool used to access the validity of clustering. Silhouette analysis helps to find the separation distance between the resulting clusters. The silhouette plot displays a measure of how close each point in one cluster is to points in the neighboring clusters. The Silhouette score has a range of [-1, 1]. Thus, if the Silhouette score has a value near +1 that indicates the sample is far away from the neighboring clusters. A value of 0 indicates that the sample is on or very close to the decision boundary between two neighboring clusters and negative values indicate that those samples might have been assigned to the wrong cluster.

#### 4.2.2. Algorithm (Silhouette Score for each Data Object)[6]

(i)      Find the mean distance of the point (i) with respect to all other points in the cluster it is assigned (say a). Let it be named as a(i).
(ii)     Find the mean distance of the point (i) with respect to its closest neighboring cluster (say b). Let it be named as b(i).
(iii)    Silhouette score for particular data point,
               $S[i] = [b(i)-a(i)] / \max[b(i),a(i)]$

For the Silhouette score, S[i] to be close to 1, a(i) has to be very small compared to b(i), that is a(i)<<<b(i). The average Silhouette score can be calculated by taking the mean of silhouette scores of all samples in the dataset. The silhouette score ranges from −1 to +1, larger values represent that the data object is very similar to its own cluster and dissimilar to neighboring clusters. These metrics are independent of the absolute values of the labels; thus a permutation of the class or cluster label values will not change the score values in any way.

### 4.2.3. Within Sum of Squares (WSS)

The WSS (Within Sum of Squares) is used to estimate the number of clusters. If k is the number of clusters then as 'k' increases, the sum of squared distance tends to zero. i.e., if we set 'k' to its maximum value n (where n is the number of samples) then each sample will form its own cluster thus the sum of squared distances will be equal to zero. In this paper, we use the K-means inertia plot to validate the number of clusters based on the distortion/ WSS.

WSS measures the squared average distance of all the points within a cluster to the cluster centroid. To calculate WSS, first find the Euclidean distance between a data point and the centroid to which it is assigned. Then iterate this process for all data points in that cluster, and then sum the values for the cluster and divide by the number of data points. Finally, you calculate the average across all clusters to get the average WCSS.

## 5. PROPOSED METHODOLOGY

For a given labeled dataset, perform the feature scaling and figure out which one gives better scores in terms of WSS. Perform clustering using those normalized datasets and calculate the score using external and internal evaluation methods. Then select the one that has high scores for all the metrics. Finally, cluster analysis is performed by means of the frequency distribution. A frequency distribution is a tabular representation of data showing the outcomes in named classes. Thus this analysis helps in comparing the clustered result with the actual classes in the dataset. This whole process is then repeated for hierarchical agglomerative clustering as well. The graphical tool (dendrogram) in the case of hierarchical clustering helped to get insight into the number of cluster solution and that was compared with the k-means inertia plot to have a double validation on the number of clusters chosen. The proposed methodology helps in validating the number of clusters and also in evaluating clustering results.

## 6. EXPERIMENTAL SETUP AND RESULTS

For the experiment, we used the famous IRIS dataset. Python programming was used for both the clustering techniques. Initially, the dataset was split into 80-20 for training purposes. The number of records given for training the model was 120 samples with 4 features. As the IRIS dataset is labeled, the number of unique values in the training data after sampling, but before clustering is given below

```
Unique Values : ['Iris-setosa' 'Iris-versicolor' 'Iris-virginica']
Occurrence Count : [43 38 39]
Unique Values along with Count in the training data
Iris-setosa  Occurs :  43  times
Iris-versicolor  Occurs :  38  times
Iris-virginica  Occurs :  39  times
```

Figure 1: Unique Values in the Training data

Table 1: Scores obtained based on K-means after performing various feature scaling on the dataset.

| | K-means | | | |
|---|---|---|---|---|
| | Inertia | Homogeneity | Completeness | V-measure |
| Before Scaling | 65.17 | - | - | - |
| Applying Standardization | 106.5 | - | - | - |
| Applying PCA | 53.89 | - | - | - |
| Applying MaxAbs | 21.58 | 0.94 | 0.94 | 0.94 |
| Applying MinMax | 54 | - | - | - |
| Normalizer | 0.24 | 0.93 | 0.94 | 0.94 |

The K-means clustering was performed on normalized and non-normalized IRIS datasets using various methods such as standardization, minmax, MaxAbs, normalizer and also by dimensionality reduction. K-means, when applied after normalizing the IRIS dataset gave the best score in terms of Homogeneity, V-measure, and Completeness.

The homogeneity, V-measure, and completeness were calculated only for those 2 cases (MaxAbs and Normalizer) were the distortion is minimum and out of those, the normalized dataset was chosen for further analysis. 'Normalization based K-means Clustering algorithm [4]', has proved that applying normalization before performing clustering will improve the execution time and speed.

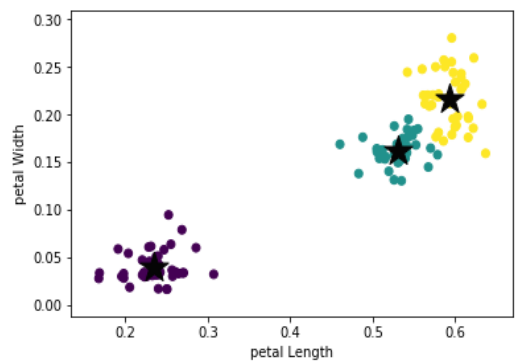

Figure 2: Scatter plot after applying K-means on normalized data.

```
Species  Iris-setosa  Iris-versicolor  Iris-virginica
labels
0           43             0                0
1           0              36               0
2           0              2                39
```

Figure 3: Frequency distribution of labels after K-means clustering on training data

Eventhough for labeled data we know the number of classes in prior, we have compared the result of elbow method with that of a dendrogram using 'ward' as the linkage method for a

Computer Science & Information Technology (CS & IT)    43

second level of validation on the number of clusters, which can be used as a method in the case of unlabeled data for future work.

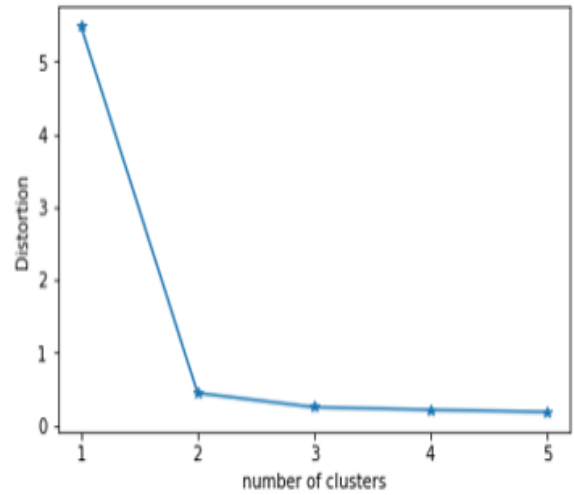

Figure 4: K-means inertia plot.

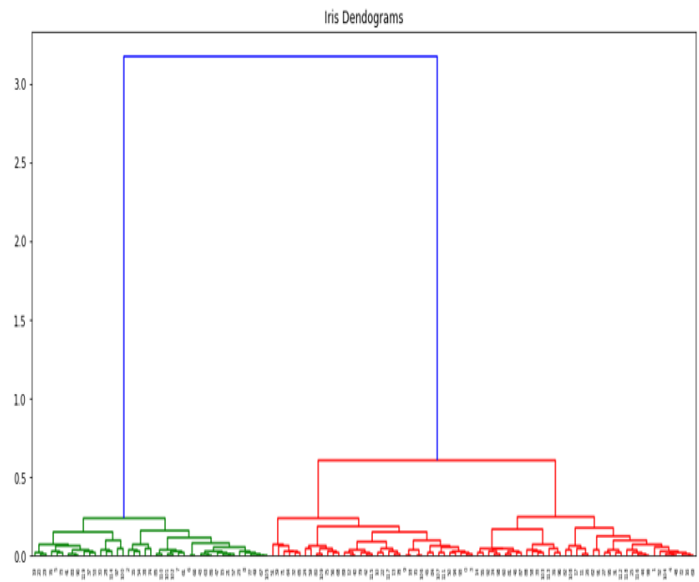

Figure 5: Iris Dendrogram.

Using the dendrogram we have taken the cut off at distance 0.5, as the jump in the distance is pretty obvious. If we draw a horizontal line at 0.5 then we end up with three clusters. On the x-axis, it provides you with the details on which clusters get merged at various distances. Thus comparing both the elbow and dendrogram we can conclude that the number of clusters for this dataset is 3 which is true based on the labels. In the case of Hierarchical clustering on the IRIS dataset, we tried different linkage methods and the highest score was obtained when the linkage method used was 'average linkage'.

Comparing different hierarchical linkage methods on the IRIS dataset with cluster count as three produced the following results.

Table:2 Hierarchical Clustering on Normalized IRIS dataset.

| Hierarchical Clustering on Normalized IRIS dataset | | | |
|---|---|---|---|
| Linkage methods Used : | Homogeneity | Completeness | V-measure |
| Ward | 0.88 | 0.885 | 0.884 |
| Average | 0.93 | 0.93 | 0.93 |
| Complete | 0.90 | 0.91 | 0.90 |
| Single | 0.59 | 0.94 | 0.73 |

```
  Species  Iris-setosa  Iris-versicolor  Iris-virginica
labels
0                    0                1              38
1                   43                0               0
2                    0               37               1
```

Figure 6: Frequency distribution table after applying average as the linkage method

| No of Clusters | Average Silhouette Score |
|---|---|
| 2 | 0.67 |
| 3 | 0.55 |
| 4 | 0.50 |
| 5 | 0.49 |
| 6 | 0.39 |

Table 3: Average Silhouette score based on k-means when cluster count =2,3,4,5,6.

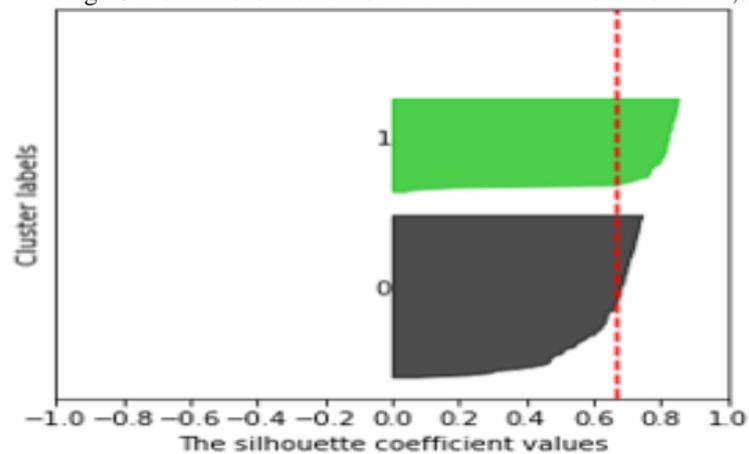

Figure 7: Silhouette plot for K-means clustering on the IRIS dataset when cluster count is 2.

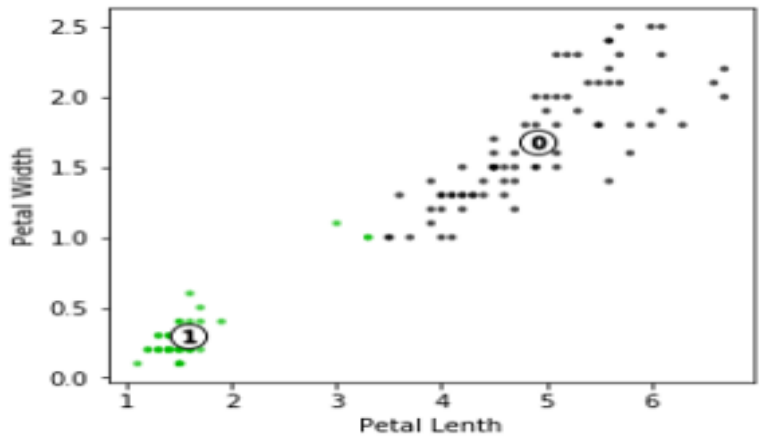

Figure 8: Visualizing the clustered IRIS data when cluster count is 2.

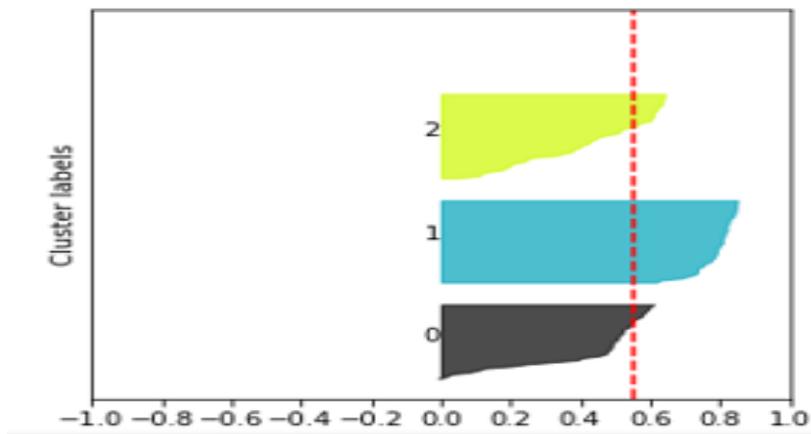

Figure 9: Silhouette plot for K-means clustering on the IRIS dataset when cluster count is 3.

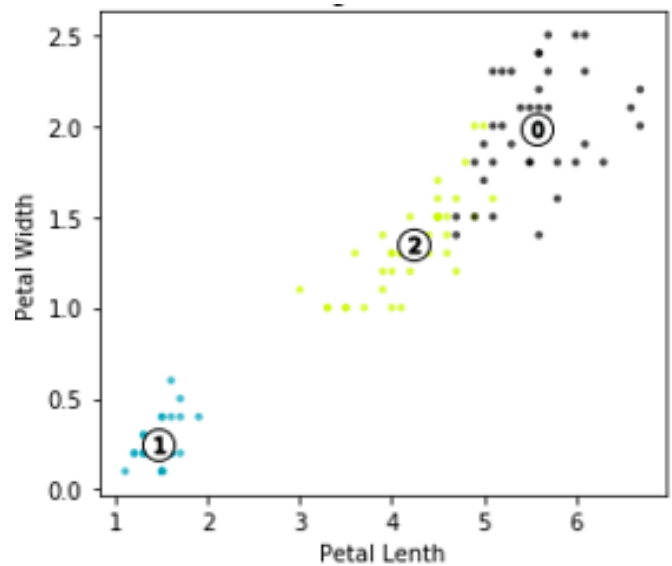

Figure 10: Visualizing the clustered IRIS data when cluster count is 3.

The figure (7,9) shows the silhouette plot for when cluster count is 2 and 3 and the figure (8,10) shows the clustered data points when cluster count is 2 and 3. Even though the average silhouette score is highest when cluster count is 2, we choose the number of clusters as 3 since the silhouette plot for cluster 0 when cluster count is 2 is bigger in size owing to the grouping of the 2 sub-clusters into one big cluster. However, when cluster count is 3, all the plots are more or less of similar thickness and hence are of almost similar sizes as can be verified from the labeled scatter plot. Also when cluster count is 3, the plot for each cluster is above the mean value (0.55) represented by a dotted line, and the width of the plot is uniform as possible. This shows that on average each datapoint is more similar to the points on its own cluster. Thus we can conclude that the number of clusters for this IRIS dataset has to be 3 to get better cluster results in terms of accuracy while performing K-means and Hierarchical Clustering.

## 7. CONCLUSION

In general, in this paper, we cross-checked the result of the dendrogram from the hierarchical agglomerative clustering and elbow method in K-means clustering to validate the best number of clusters as in real world most of the datasets come without true labels. Then to evaluate the quality of clustering based on different feature scaling and setting the number of clusters as 3, we calculated the V-measure, completeness and homogeneity score. These external validation methods were used for evaluating the quality of clustering. As a final step, we quantify the correctness of the partition by comparing the frequency distribution in the classes in the original training data with the clustered data. As an internal evaluation method, Silhouette analysis and WSS were also done on the training data. All the results are consistent in the sense that Silhouette and WSS present an appropriate number of clusters at the same k-value with regard to maximum average silhouette value and knee point in the case of WSS for the IRIS dataset. Also, we proved from the frequency distribution table that using k-means++ as the initialization procedure for the k-means, improved the clustering results using a clever seeding of the initial cluster centroids. The frequency table shows that the quality of the clustered classes was better while using k-means++ for cluster initialization than the result obtained while using K- medoids in paper [7] for the same IRIS dataset.

## 8. FUTURE WORK

As future work, we will use other labeled datasets to validate our methodology. Those datasets will be of different sizes to make sure that our method works well on all different datasets.

## AUTHORS

**Anupriya Vysala** is a Graduate Student in the Department of Compruter Science at Bowie State University. She has her MS degree in Computer Science from George Washington University.

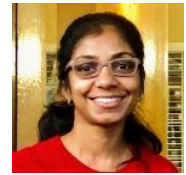

**Dr. Joseph Gomes** is an Associate Professor in the Department of Computer Science at Bowie State University. He has his MS and DSc degree in Computer Science from George Washington University.

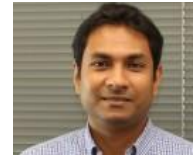